\documentclass[sigconf,nonacm]{acmart}     
\AtBeginDocument{%
  \providecommand\BibTeX{{%
    \normalfont B\kern-0.5em{\scshape i\kern-0.25em b}\kern-0.8em\TeX}}}

\usepackage{amsmath}

\usepackage{multirow}
\usepackage{subcaption}
\usepackage{graphicx}

\usepackage{array}
\newcolumntype{H}{>{\setbox0=\hbox\bgroup}c<{\egroup}@{}}

\usepackage[acronym]{glossaries}
\newacronym{MODEL}{SCGC}{Self-Supervised Contrastive Graph Clustering}  
\newacronym{GNN}{GNN}{Graph Neural Network}
\newacronym{GAT}{GAT}{Graph Attention Network}
\newacronym{GCN}{GCN}{Graph Convolutional Network}
\newacronym{CNN}{CNN}{Convolutional Neural Network}

\begin{document}

\title{SCGC : Self-Supervised Contrastive Graph Clustering}


\author{Gayan K. Kulatilleke}
\authornotemark[1]
\email{g.kulatilleke@uqconnect.edu.au}
\affiliation{
    \institution{University of Queensland}
    \city{Brisbane}
    \country{Australia}
}
\author{Marius Portmann}
\email{marius@itee.uq.edu.au}
\affiliation{
    \institution{University of Queensland}
    \city{Brisbane}
    \country{Australia}
}
\author{Shekhar S. Chandra}
\email{shekhar.chandra@uq.edu.au}
\affiliation{
    \institution{University of Queensland}
    \city{Brisbane}
    \country{Australia}
}
\authornote{\authornotemark[1]Corresponding author.}


\begin{abstract} 
Graph clustering discovers groups or communities within networks. Deep learning methods such as autoencoders (AE) extract effective clustering and downstream representations but cannot incorporate rich structural information. While Graph Neural Networks (GNN) have shown great success in encoding graph structure, typical GNNs based on convolution or attention variants suffer from over-smoothing, noise, heterophily, are computationally expensive and typically require the complete graph being present.
Instead, we propose \acrfull{MODEL}, which imposes graph-structure via contrastive loss signals to learn discriminative node representations and iteratively refined soft cluster labels. We also propose \acrshort{MODEL}*, with a more effective, novel, \textbf{I}nfluence \textbf{A}ugmented \textbf{C}ontrastive (IAC) loss to fuse richer structural information, and half the original model parameters. 
\acrshort{MODEL}(*) is faster with simple linear units, completely eliminate convolutions and attention of traditional GNNs, yet efficiently incorporates structure. It is impervious to layer depth and robust to over-smoothing, incorrect edges and heterophily. It is scalable by batching, a limitation in many prior GNN models, and trivially parallelizable. We obtain significant improvements over state-of-the-art on a wide range of benchmark graph datasets, including images, sensor data, text, and citation networks efficiently. Specifically, 20\% on ARI and 18\% on NMI for DBLP; overall 55\% reduction in training time and overall, 81\% reduction on inference time.

Our code is available at : https://github.com/gayanku/SCGC
\end{abstract}


\maketitle

\section{Introduction}
Research into graphs has been receiving increased attention due to the high expressiveness and pervasiveness of graph structured data \cite{kulatilleke2021fdgatii}. Its unique non-Euclidean data structure is ideally suited to represent diverse feature rich domains for machine learning \cite{wang2019attributed}: 
\citet{chen2018measurement} carried out deep analysis of social forum interactions for node classification; \citet{kipf2016semi} predicted Facebook friend suggestions and \citet{samtani2017exploring} analysed dark web social network forums to obtain cyber threat intelligence. 

Graph clustering discovers groups or communities within networks by partitioning similar nodes into disjoint groups \cite{bo2020structural, wang2019attributed}. Clustering has been used for images \cite{zhong2021graph,hu2021graph}, text \cite{pan2018adversarially, bo2020structural} and social networks \cite{samtani2017exploring,chen2018measurement}. To date, deep clustering methods, based on Auto Encoders (AE) \cite{hinton2006reducing,xie2016unsupervised, guo2017improved, bo2020structural, peng2021attention} have achieved state-of-the-art performance. In order to exploit the rich information present in the structure, many researchers \cite{kipf2016variational,wang2019attributed,pan2018adversarially,bo2020structural, peng2021attention} have combined \acrfull{GNN} variants with AEs \cite{wu2020comprehensive}.

Although these models achieved remarkable improvements, and state-of-the-art in clustering, the reliance on \acrshort{GNN} for structure incorporation is challenging due to (a) over-smoothing, (b) noisy neighbours (heterophily), and (c) the suspended animation problem \cite{kulatilleke2021fdgatii}. To facilitate interaction between nodes that are not directly connected, a \acrshort{GNN} stacks layers \cite{kipf2016semi} which leads to over-smoothing where node representations become indistinguishable due to too much mixing \cite{wu2019simplifying}. To alliviate this, most GNNs are shallow and cannot benefit from deep models. Models such as GCNII \cite{chen2020simple} and FDGATII \cite{kulatilleke2021fdgatii} are able to achieve higher depths but still require appropriate depth to be pre-determined and use computationally expensive convolutions or softmax attention operations.

Recently there has been a shift towards more simpler and efficient model implementations \cite{wu2019simplifying,maurya2021improving,tolstikhin2021mlp,hu2021graph, kipf2019contrastive}.
It is well known that the structural information represents the underlying dependencies among nodes \cite{bo2020structural}. However, such dependencies can be direct (local or first-order structure) or indirect (long term) dependencies of one or multiple orders in arbitrary compositions. Thus, it is non-trivial to model such an unpredictable latent structure with a fixed pre known convolutional or other layer structure. 

While prior work \cite{hu2021graph, you2020graph, kipf2019contrastive} has used contrastive loss as a means to guide embeddings, these works either use augmented images or graphs, still contain a form of \acrshort{GNN}, or require supervision. To the best of our knowledge, there are no models that can perform self-supervised clustering on graphs without using a \acrshort{GNN}.

In this work, we propose a novel deep clustering method, \acrfull{MODEL}, which uses contrastive loss on the end embeddings, rather than attempting to match the latent node dependency dynamics, as a means to enforce graph structure and guide the optimization. This completely eliminates convolutions and the need to carry adjacency information through the model, decoupling the model structure from the latent node dependency structure. Thus, the \acrshort{MODEL} structure can remain unchanged across diverse data, as we demonstrate using benchmark data sets from image, text and graph modalities.
Specifically, we use an AE and impose graph structure as a contrastive loss objective. While \acrshort{MODEL} can be used with any AE variant, for comparison purposes we use the simple AE as in \cite{bo2020structural, peng2021attention}. To facilitate effective clustering, we further use a self-supervised approach based on promoting confident soft labels, to jointly guide cluster optimization.

Our \acrshort{MODEL} is computationally efficient as it consists of a simple linear layer (MLP) based AE. There are no expensive convolutions or softmax operations. Further, passing adjacency information though the model is not required. Such models can be an attractive option for edge and resource constraint applications. By only having soft structure enforcement via contrastive loss, the model is more robust to noisy edges. Further, as we are using a probability distribution based self-supervision mechanism, the model is also robust to feature noise and class/label noise, i.e. heterophily. In summary, our main contributions are:
\begin{itemize}
\item We introduce an efficient novel deep clustering model, \acrshort{MODEL}, that completely removes the necessity to carry the structure/edge information throughout the learning layers. 
To the best of our knowledge, this is the first deep model to effectively perform graph node clustering without using GNNs. 

\item We propose a novel \textbf{I}nfluence \textbf{A}ugmented \textbf{C}ontrastive (IAC) loss to incorporate graph structure, which can effectively transform any model to become graph-aware, and give theoretical insights and experimental evidence of its superiority over simple contrastive loss.

\item We propose \acrshort{MODEL}*, a leaner variant exploiting pre-learnt centroids, which uses half the original parameters t significantly boosting training and inference speeds.
\item Extensive experiments on 6 benchmark datasets show both \acrshort{MODEL} and \acrshort{MODEL}* outperform state-of-the-art graph clustering methods in accuracy as well as training and inference efficiency (even in its its un-batched implementation).
\end{itemize}

\section{Related Work}\label{RelatedWork}
\subsection{Auto encoders (AE)}
An AE \cite{hinton2006reducing} based latent embeddings learning approach can be applied to purely unsupervised environments including clustering \cite{wang2019attributed}. Early work on graph clustering relied purely on node features: \citet{hinton2006reducing} introduced the classical auto encoder; \citet{xie2016unsupervised} introduced deep embedded clustering method (DEC) that incorporated KL divergence into the auto encoder; \citet{guo2017improved} combined a reconstruction loss to improve DEC. However, the complexity of graph topological structure imposes significant challenges on clustering \cite{wang2019attributed} which AEs alone cannot solve.

\subsection{Incorporating graph structure}
A \acrshort{GNN} performs node aggregation based on the neighbourhood structure to obtain effective low dimensional embedding vectors \cite{wu2020comprehensive}. \acrshort{GNN} variants attempting to build effective and efficient models manly differ in how the aggregation and subsequent combining of the node features is done \cite{kulatilleke2021fdgatii}: \acrfull{GCN} \cite{kipf2016semi} uses convolution \cite{lecun1995convolutional}; GraphSage \cite{hamilton2017inductive} uses max-pooling and \acrfull{GAT} \cite{velivckovic2018graph} uses attention. 

In order to benefit from rich structural information, \acrshort{GNN}s have been combined with AE: \citet{kipf2016variational} proposed the graph auto encoder (GAE) and its variational variant (VGAE) by applying convolution \cite{lecun1995convolutional} to an AE \cite{hinton2006reducing}; \citet{wang2019attributed} proposed DAEGC which used graph attention \cite{velivckovic2018graph} with GAE; \citet{pan2018adversarially} proposed ARGA by introducing an adversarial regularizer to GAE. Recent work \cite{bo2020structural, peng2021attention} demonstrated benefits of decoupling the AE (features) and GNN (structure) components. Specifically, \citet{bo2020structural} proposed SDCN which coupled DEC and GCN via a fixed delivery operator and reconstructed the features rather than the adjacency matrix; \citet{peng2021attention} proposed AGCN by extending SDCN with a more flexible attention-based delivery operator and used multi scale information. Although these models achieved remarkable improvements, and state-of-the-art in clustering, they rely on GNN for structure incorporation. 

\subsection{Towards simpler graph models}
Recently there has been a shift towards more simpler and efficient model implementations: \citet{wu2019simplifying} proposed SGC by successively removing activation layers and adding a pre-computed $r^{th}$ power adjacency matrix to capture non-neighbour relations; \citet{maurya2021improving} proposed FSGNN by decoupling the node feature aggregation from depth of graph neural network by using an array of pre-computed $r$-th power adjacency metrics; MLP-Mixer  \cite{tolstikhin2021mlp}, exclusively based on multi-layer perceptron (MLP), attains competitive scores on image classification benchmarks; Graph-MLP \cite{hu2021graph} uses MLP for graph citation networks; C-SWMs \cite{kipf2019contrastive} uses MLP for compositional objects. 

\subsection{Contrastive loss}
Some work has used contrastive loss as a means to guide embeddings \cite{hu2021graph, you2020graph, kipf2019contrastive}. GraphCL \cite{you2020graph} uses contrastive learning on augmented views for GNN pre-taining; Graph-MLP \cite{hu2021graph} uses contrastive loss for graph node classification and \cite{kipf2019contrastive} used contrastive loss between successive images to learn a delta for object detection. However, these work either use augmented images or graphs, still contain a form of \acrshort{GNN}, or require supervision.

\begin{figure*}[t]
  \includegraphics[width=\textwidth]{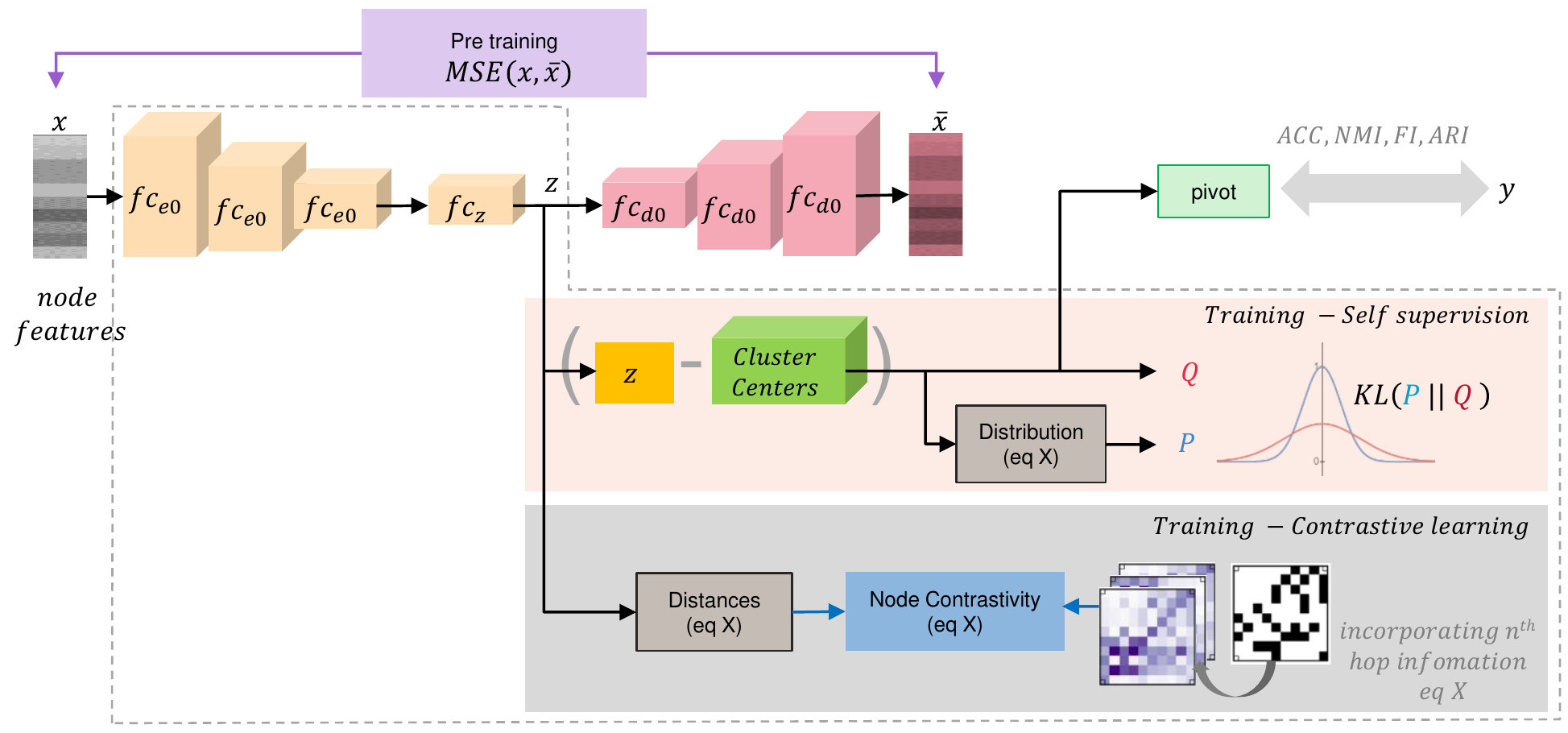}
  \caption{\acrshort{MODEL} jointly learns structure and cluster assignment via probabilistic soft assignment. Dotted section outlines the more efficient \acrshort{MODEL}* based on Influence contstration. Cluster centroids $\mu$ are obtained by pre-training the AE for reconstruction.}
  \label{fig_Model}
\end{figure*}

\section{Proposed Model}\label{ProposedModel}
Distinct from previous \acrshort{GNN} models that carry and use neighbour information as adjacency matrix or via message passing, which leads to complex structure and heavy computation \cite{hu2021graph}, we use a simple AE and apply a novel node influence based contrastive loss to superimpose graph structure as shown in Figure~\ref{fig_Model}. Next we introduce its framework and the two-phase training process, which simultaneously learns discriminative embeddings and clusters. 

\subsection{Graph structure by contrastive loss}
Contrastive loss makes positive or connected nodes closer and negative or unconnected nodes further away in the feature space. Motivated by this, we use an adjacency guided contrastive loss to incorporate graph structure into embeddings. Specifically, we fist compute the similarity or distance between two embeddings, then use augmented edge information to determine positive samples.

\subsubsection{\textbf{I}nfluence \textbf{A}ugmented \textbf{C}ontrastive (IAC) loss}
it is intuitive that edge-connected, and thus related, nodes share some similarity with those that are not \cite{hu2021graph}. Additionally, graphs benefit from the unique ability where, non-adjacent nodes at multiple depths can have arbitrary dissimilar and additive effects on a node. For example, for a given $R$ depth, we can define the total influence as:
\begin{equation}
   \gamma_{ij} = \operatorname{Effect}_{Rij}=\sum_{r=1}^{R} \alpha_{ijr} \operatorname{relationship}_r(i,j),
\label{EQ_Influence}
\end{equation}
where $\alpha_{ijr}$ is the coefficient denoting the relationship between nodes $i,j$ at depth $r$. While this has seldom been exploited, it can carry richer information. Prior \acrshort{GNN} models naively assume a fixed depth relationship, i.e. a single $r$ obtained via hyper parameter searches. Layers of a typical \acrshort{GNN} attempt to exploit this feature with a rigid layer structure. However, this layer structure needs to align with the latent influence structures. Further, \acrshort{GNN} layers are universally applied to all nodes, thus incorrectly assuming all nodes get the same influence from different depth effects (essentially the assumption that there is only one fixed $r$). As a result, \acrshort{GNN} models are often sub optimal. Further, most \acrshort{GNN} models cannot exceed 2 layers in depth due to over smoothing, which models such as FDGATII \cite{kulatilleke2021fdgatii} and GCNII \cite{chen2020simple} attempts to solve.

Given we know $\gamma_{ij}$, we formulate IAC loss for the $i^{th}$ node as: 
\begin{equation}
\ell_{i}=-\log \frac
{\sum_{j=1}^{B} \mathbf{1}_{[j \neq i]}   \gamma_{ij}  \exp \left(\operatorname{distance}\left(\boldsymbol{z}_{i}, \boldsymbol{z}_{j}\right) / \tau\right)}
{ 10^{-8} + \sum_{k=1}^{B} \mathbf{1}_{[k \neq i]} \exp \left(\operatorname{distance}\left(\boldsymbol{z}_{i}, \boldsymbol{z}_{k}\right) / \tau\right)} ,
\label{EQ_IC_DEF}
\end{equation}
where $\tau$ denotes the temperature parameter and $\gamma_{ij}$ is the influence of the connection between node $i$ and $j$.

Essentially, for each node, its \textit{cumulative} $R$-hop neighbour influence is used to distinguish positive samples, which we contrast with all nodes. IAC loss encourages influential nodes to be closer than the non-influential nodes in the embedding space. Next, we outline how cumulative influence can be computed.


\subsubsection{Determining Influence} 
As real-world graphs are usually extremely sparse, most of the entries in the adjacency matrix $A$ are zero \cite{chen2019fast}. However, absence of an edge between two nodes $i$ and $j$ does not imply no association; there can still be strong associations, i.e.: high-order proximities. This intuition motivates exploiting higher-order relationships in the graph, which is typically performed by raising $A$ to r-th power \cite{hu2021graph, chen2019fast}. The $ij$-th entry of $A^r$ gives the number of $r$-length walks.

Similarly, for the normalized adjacency matrix $\widehat{A}$:
\begin{equation}
\widehat{A}=\mathbf{D}^{-0.5} \left(A + I\right) \mathbf{D}^{-0.5},
\end{equation}
where $I$ is the self-connection and $\mathbf{D}$ is the diagonal matrix with $\mathbf{D}_{ij}=\sum_{j} \widetilde{A}_{ij}$. The $r$-th power provides the strength of the $r$-th hop relationship between nodes $i$ and $j$ \cite{hu2021graph}.

We compute the influence as an additive form of compositional node relationships, rather than limit to some arbitrary $r$-th hop neighbourhood. Specifically, we define the $R$-th cumulative power of the normalized adjacency matrix as $\widehat{A}^{R}$: $\gamma_{ij} = \widehat{A}^{R}_{ij}$ where $\widehat{A}^{R}= \sum_{r=1}^{K} \widehat{A}^{r}$. Importantly, $\widehat{A}^{R}$ contains the aggregated set of \textit{all} previous neighbourhood hops relationships from $k= 1 \cdots K$. Computing $\widehat{A}^{K}$ needs only be done once, prior to training, using the adjacency matrix, adding very little overheads. It is noted that $\gamma^{ij}$ gets non-zero values only if node $j$ has some non-zero influence from its $r$-hop neighbour of node $i$. 

\begin{equation*}
     \gamma_{ij} \begin{cases}
= 0,& \parbox[t]{5.5cm}{node $i$ has no influence, nor is it connected to node $j$ for $K$ hops} \\
\ne 0,& \parbox[t]{5.5cm}{node $i$'s cumulative influence from $j$ within an $R$-hop neighbourhood}\\
\end{cases}
\label{EQ_IL}
\end{equation*}

Distinct from our work on influence, \citet{hu2021graph} proposed cosine similarity based NContrast (NC) loss for classification, where for each node, only the $r$-th hop neighbourhood is considered, not the fuller additive influence. We adopt \cite{hu2021graph} to self-supervised clustering using Equation~\ref{EQ_IC_DEF} and: 
\begin{equation*}
     \gamma_{ij} \begin{cases}
= 0,& \text{node $j$ is the $r$-hop neighbour of node $i$} \\
\ne 0,& \text{node $j$ is not the $r$-hop neighbour of node $i$ }\\
\end{cases}
\label{EQ_NC}
\end{equation*}

The complete contrastive loss, for IAC or NC, is defined as: 
\begin{equation}
    loss_{contrastive}=\frac{1}{B}\sum_{i=1}^{B}\ell_{i}
\label{EQ_LOSSIC}
\end{equation}

\subsection{Self supervised clustering}
Graph clustering is essentially an unsupervised task with no feedback available to guide the optimization progress which makes it challenging. To this end, we use probability distribution derived soft-labels as a self-supervision mechanism for cluster enhancement, which effectively superimposes clustering on the embeddings.

Similar to existing work \cite{wang2019attributed, guo2017improved,xie2016unsupervised}, we first obtain the soft cluster assignments probabilities $q_{iu}$, for embedding $z_i$ and cluster centre $\mu_u$, using the student's $t$-distribution \cite{maaten2008visualizing} as a kernel to measure the similarity between the embedding and centroid, in order to handle differently scaled clusters and be computationally convenient \cite{wang2019attributed} as follows:
\begin{equation} 
q_{iu} = \frac{(1+\left\|z_i - {\mu}_u\right\|^2/\eta)^{-\frac{\eta+1}{2}}}
        {\sum_{u'}(1+\left\|z_i - {\mu}_{u'}\right\|^{2}/\eta)^{-\frac{\eta+1}{2}}},
\label{eq-QIJ}
\end{equation} 
where, cluster centres $\mu$ are initialized by $K$-means on embeddings from the pre-trained AE and $\eta$ is the Student’s $t$-distribution's degree of freedom. We use $Q=[q_{iu}]$ as the distribution of the cluster assignments of all samples and keep $\eta$=1 for all experiments as in prior work \citep{bo2020structural,peng2021attention}

Nodes closer to a cluster centre have higher soft assignment probabilities in $Q$. By raising $Q$ to the second power and normalizing, we define a target distribution $P$ that emphasises the \textit{confident} assignments, which is defined as: 
\begin{equation} 
p_{iu} = \frac{q_{iu}^2/\sum_i q_{iu}}{\sum_k{(q_{ik}^2/\sum_i q_{ik})}},
\label{eq-PIJ}
\end{equation} 
where $\sum_{i}q_{iu}$ is the soft cluster frequency of centroid $u$.

In order to make the data representation closer to cluster centres and improve cluster cohesion, we minimise the KL divergence loss between $Q$ and $P$ distributions, which forces the current distribution $Q$ to approach the more confident target distribution $P$. We self-supervise cluster assignments\footnotemark[1] by using distribution $Q$ to target distribution $P$, which then supervises the distribution $Q$ in turn by minimizing the KL divergence as:  
\begin{equation} 
loss_{cluster}= KL(P||Q)=\sum_{i}\sum_{u}p_{iu}log\frac{p_{iu}}{q_{iu}},
\label{eq-KLPQ}    
\end{equation} 

\footnotetext[1]{We follow \cite{bo2020structural}, which uses the term 'self-supervised' to be consistent with the GCN training method.}

KL divergence updates  models more gently and lessens severe disturbances on the embeddings \cite{bo2020structural}. Further, it can accommodate both the structural and feature optimization targets of \acrshort{MODEL}.

\subsection{Initial centroids and embeddings}
In order to extract the node features and obtain the initial embeddings $z$ and cluster centroids $\mu$ for optimization, we use an AE based pre-training phase. First, we use the encoder-decoder to extract latent embeddings $z$ by minimizing the reconstruction loss between the raw data $\mathbf{X}\in\mathbb{R}^{n\times d}$ and the reconstructed data $\hat{\mathbf{X}}\in\mathbb{R}^{n\times d}$, i.e.,
\begin{equation} 
\begin{aligned}
    & loss_{recon} = \left\| \mathbf{X} - \hat{\mathbf{X}} \right\|^2_F\\
    & \rm{encoder:} \quad H_{enc}^k=\phi\left(W_{enc}^k H_{enc}^{k-1} + b_{enc}^k\right),\\
    & \rm{decoder:} \quad \hat{H}_{dec}^k=\phi\left(W_{dec}^k \hat{H}_{dec}^{k-1} + b_{dec}^k\right),\\
    & \rm{where} \quad \emph{k}={1,\cdots,\emph{K} },\\
    & \rm{and} \quad H_{enc}^0 = X, \quad \hat{H}_{dec}^0 = H_{enc}^K, \quad \hat{X} = H_{dec}^K.
\end{aligned}
\label{eq-AE}
\end{equation} 
On completion of the pre training, we obtain $Z = \hat{H}_{dec}^0 = H_{enc}^K$ and use K-means to obtain the cluster centres $\mu$.

\subsection{Final proposed models}
Our preliminary experiments showed that, once quality centroids $\mu$ are available, the feature reconstruction objective can be made redundant. Thus, our AE can then simply become an MLP similar to the encoder component in Equation~\ref{eq-AE}, effectively halving the AE based parameters and training effort.
Thus, we propose two model variants \acrshort{MODEL} and \acrshort{MODEL}* as,
\begin{equation} 
\begin{aligned}
    & \rm{\acrshort{MODEL}:} \quad \L_{final} =  \alpha loss_{nc}(K,\tau) + \beta loss_{cluster} + loss_{recon},\\
    & \rm{\acrshort{MODEL}^*:}\quad \L_{final} =  \alpha loss_{iac}(K,\tau) + \beta loss_{cluster},\\
\end{aligned}
\label{eq-LOSSES}
\end{equation} 
where $\alpha>0$ is the hyper-parameter that balances structure incorporation and $\beta>0$ controls the cluster optimization.  

\subsection{Complexity Analysis} \label{Complexity}
Given the input data dimension $d$ and dimensions of the AE layers as $d_{1}, d_{2}, \cdots, d_{L}$, following \cite{bo2020structural}, the size of weight matrix of the first encoder layer is $W^1_{enc} \in \mathbb{R}^{d \times d_{1}}$. With an input data size $N$, the time complexity is $O_1=\mathcal{O}(Nd^{2}d^{2}_{1}...d^{2}_{L})$ for \acrshort{MODEL}-AE and $O^*_1=\mathcal{O}(Nd^{2}d^{2}_{1}...d^{2}_{L/2})$ for \acrshort{MODEL}*-AE. Assuming $K$ clusters, from Equation~\ref{eq-QIJ}, the time complexity is $O_2 = \mathcal{O}(NK+N\log N)$ following \cite{xie2016unsupervised}. 

For the contrastive loss, we compute $\|z\|_{2}^{2}$ and $z_{i}\cdot z_{j}$ for all $N$. Thus the time complexity is $O_3 = \mathcal{O}(NNd_z)$ where $d_z$ is the embedding dimension. While this results in a theoretical time complexity of $\mathcal{O}(N^2)$, given that $Z\cdot Z^T$ is symmetrical, we only need to compute half of the result, and as the transpose of a matrix is the same matrix with indices swapped, caching can have over twice the impact. \cite{chen2019fast} presents an algorithm to obtain a similarity matrix that preserves graph transitive relationships which is formulated as matrix chain multiplications, so that applying random projection costs linear time. Lastly, batching (see Section~\ref{Implementation}) allows the use of $b<<N$. We experimentally show that the efficiency of our approach, with none of the above  optimizations, is still competitive for average data sets, thus implying batching alone is sufficient for scalability.

\section{Experiments}\label{Experiment}
\subsection{Datasets}

\begin{table}
\caption{Statistics of the node clustering datasets}
\label{TABLE_datasets}
\begin{tabular}{l|c|c|c|c}
\toprule
Dataset  & Type   & Samples & Classes & Dimension \\ \midrule
USPS     & Image  & 9298    & 10      & 256       \\
HHAR     & Record & 10299   & 6       & 561       \\
Reuters  & Text   & 10000   & 4       & 2000      \\
ACM      & Graph  & 3025    & 3       & 1870      \\
CiteSeer & Graph  & 3327    & 6       & 3703      \\ 
DBLP     & Graph  & 4057    & 4       & 334       \\ \bottomrule
\end{tabular}
\vspace{-4mm} 
\end{table}
\begin{figure}
    \centering \includegraphics[width=1.00\columnwidth]{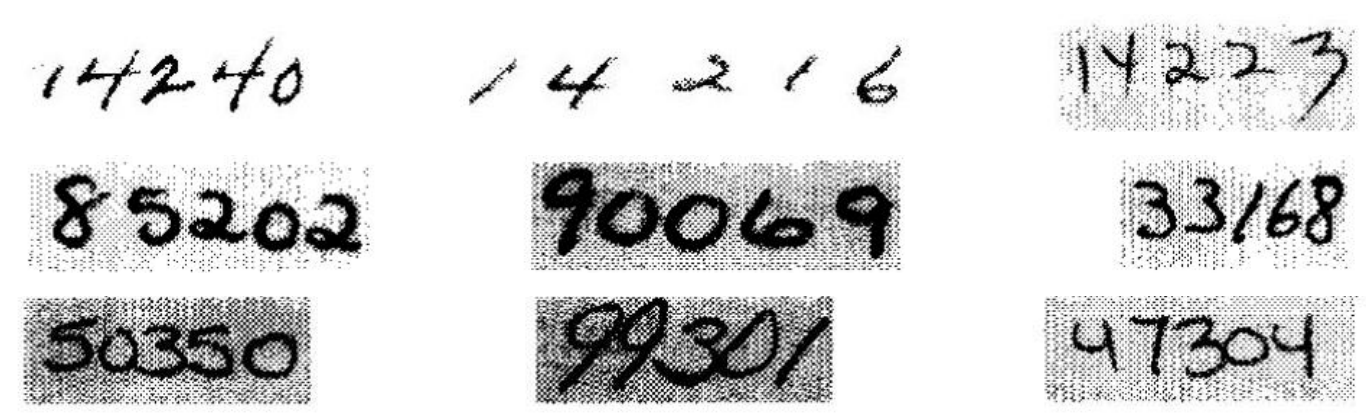}
    \caption{Sample of the USPS handwritten texts \cite{le1990handwritten}}
    \label{FIG_USPS}
\end{figure}

Experiments are conducted on six common clustering benchmarks, which includes one image dataset (USPS \cite{le1990handwritten}), one sensor data dataset (HHAR \cite{stisen2015smart}), one text dataset (Reuters \cite{lewis2004rcv1}) and three citation graphs (ACM\footnotemark[2], CiteSeer\footnotemark[4], and  DBLP\footnotemark[3]) following \cite{bo2020structural, peng2021attention}. For the non-graph data, we use undirected $\emph{k}$-nearest neighbor (KNN \cite{altman1992introduction}) to generate adjacency matrix $\mathbf{A}$ as in \cite{bo2020structural, peng2021attention}. Table~\ref{TABLE_datasets} summarizes the datasets.

\begin{itemize}
\item \textbf{USPS}\cite{le1990handwritten}, the United States Postal Service database, contains a ten class (i.e., `0'–`9') subset of 9298 grey-scale handwritten 16x16 pixels digits (Figure~\ref{FIG_USPS}) normalized to [0, 2]. 
\item \textbf{HHAR}\cite{stisen2015smart}, the Heterogeneity Human Activity Recognition dataset, contains 10299 sensor records from smart devices (i.e., phones, watches), partitioned into 6 categories of human activities : biking, sitting, standing, walking, stair-up and stair-down.
\item \textbf{Reuters}\cite{lewis2004rcv1} is a collection of English news, from which four categories (corporate/industrial, government/social, markets and economics) have been sampled for clustering. \item \textbf{ACM\footnotemark[2]} is a paper network from the ACM digital library. Edges connect papers from same author. The features are selected from KDD, SIGMOD, SIGCOMM, MobiCOMM keywords. There are three classes (i.e., database, wireless communication, data mining) by author research area.
\footnotetext[2]{http://dl.acm.org/}
\item \textbf{DBLP\footnotemark[3]} is an author network from the dblp computer science bibliography. An edge connects authors if they have a co-author relationship. Author features are bag-of-words of keywords. Authors belong to four research areas: database, data mining, machine learning, and information retrieval.
\footnotetext[3]{https://dblp.uni-trier.de}
\item \textbf{Citeseer\footnotemark[4]} is a citation network with sparse bag-of-words feature vectors for each document and a list of citation links, categorized in to six areas: agents, artificial intelligence, database, information retrieval, machine language, and HCI.
\footnotetext[4]{http://citeseerx.ist.psu.edu/index}
\end{itemize}

\begin{table*}[h]
\small
\caption{Clustering performance on six datasets (mean$\pm$std). Best results are \textbf{bold}; second best is \underline{underlineed} if it is not a \acrshort{MODEL} variant. SDCN-Q variant results are in \textit{italics}. Results reproduced from \cite{bo2020structural, peng2021attention}. \acrshort{MODEL} uses $r$-hop neighbour information. \acrshort{MODEL}* uses the novel $r$-hop cumulative Influence contrastive loss}
\label{table_results}
\setlength{\tabcolsep}{1.48mm}{
\begin{tabular}{l|c|cccccHccc|cc} 
\hline
Dataset                   & Metric & $K$-means      & AE             & DEC            & IDEC           & GAE            & VGAE           & DAEGC          & SDCN                    & AGCN                        & \acrshort{MODEL}  & \acrshort{MODEL}*  \\ \hline
\multirow{4}{*}{USPS}     & ACC    & 66.82$\pm$0.04 & 71.04$\pm$0.03 & 73.31$\pm$0.17 & 76.22$\pm$0.12 & 63.10$\pm$0.33 & 56.19$\pm$0.72 & 73.55$\pm$0.40 & 78.08$\pm$0.19          & \rm{80.98$\pm$0.28}  &  82.90$\pm$0.08      &  \textbf{84.91$\pm$0.06}\\
                          & NMI    & 62.63$\pm$0.05 & 67.53$\pm$0.03 & 70.58$\pm$0.25 & 75.56$\pm$0.06 & 60.69$\pm$0.58 & 51.08$\pm$0.37 & 71.12$\pm$0.24 & 79.51$\pm$0.27          & \rm{79.64$\pm$0.32}  &  82.51$\pm$0.07      &  \textbf{84.16$\pm$0.10}\\
                          & ARI    & 54.55$\pm$0.06 & 58.83$\pm$0.05 & 63.70$\pm$0.27 & 67.86$\pm$0.12 & 50.30$\pm$0.55 & 40.96$\pm$0.59 & 63.33$\pm$0.34 & 71.84$\pm$0.24          & \rm{73.61$\pm$0.43}  &  76.48$\pm$0.11      &  \textbf{79.50$\pm$0.06}\\
                          & F1     & 64.78$\pm$0.03 & 69.74$\pm$0.03 & 71.82$\pm$0.21 & 74.63$\pm$0.10 & 61.84$\pm$0.43 & 53.63$\pm$1.05 & 72.45$\pm$0.49 & 76.98$\pm$0.18          & \rm{77.61$\pm$0.38}  &  80.06$\pm$0.05      &  \textbf{81.54$\pm$0.06}\\ \hline
						  
\multirow{4}{*}{HHAR}     & ACC    & 59.98$\pm$0.02 & 68.69$\pm$0.31 & 69.39$\pm$0.25 & 71.05$\pm$0.36 & 62.33$\pm$1.01 & 71.30$\pm$0.36 & 76.51$\pm$2.19 & 84.26$\pm$0.17          & \rm{88.11$\pm$0.43}  &  \textbf{89.49$\pm$0.22}     &  89.36$\pm$0.16 \\
                          & NMI    & 58.86$\pm$0.01 & 71.42$\pm$0.97 & 72.91$\pm$0.39 & 74.19$\pm$0.39 & 55.06$\pm$1.39 & 62.95$\pm$0.36 & 69.10$\pm$2.28 & 79.90$\pm$0.09          & \rm{82.44$\pm$0.62}  &  84.24$\pm$0.29      &  \textbf{84.50$\pm$0.41}\\
                          & ARI    & 46.09$\pm$0.02 & 60.36$\pm$0.88 & 61.25$\pm$0.51 & 62.83$\pm$0.45 & 42.63$\pm$1.63 & 51.47$\pm$0.73 & 60.38$\pm$2.15 & 72.84$\pm$0.09          & \rm{77.07$\pm$0.66}  &  \textbf{79.28$\pm$0.28}     &  79.11$\pm$0.18 \\
                          & F1     & 58.33$\pm$0.03 & 66.36$\pm$0.34 & 67.29$\pm$0.29 & 68.63$\pm$0.33 & 62.64$\pm$0.97 & 71.55$\pm$0.29 & 76.89$\pm$2.18 & 82.58$\pm$0.08          & \rm{88.00$\pm$0.53}  &  \textbf{89.59$\pm$0.23}     &  89.48$\pm$0.17 \\ \hline

\multirow{4}{*}{Reuters}  & ACC    & 54.04$\pm$0.01 & 74.90$\pm$0.21 & 73.58$\pm$0.13 & 75.43$\pm$0.14 & 54.40$\pm$0.27 & 60.85$\pm$0.23 & 65.50$\pm$0.13 & \textit{79.30$\pm$0.11} & \rm{79.30$\pm$1.07}  &  \textbf{80.32$\pm$0.04}     &  79.35$\pm$0.00 \\
                          & NMI    & 41.54$\pm$0.51 & 49.69$\pm$0.29 & 47.50$\pm$0.34 & 50.28$\pm$0.17 & 25.92$\pm$0.41 & 25.51$\pm$0.22 & 30.55$\pm$0.29 & \underline{\textit{56.89$\pm$0.27}} & \textbf{57.83$\pm$1.01}  &  55.63$\pm$0.05      &  55.16$\pm$0.01 \\
                          & ARI    & 27.95$\pm$0.38 & 49.55$\pm$0.37 & 48.44$\pm$0.14 & 51.26$\pm$0.21 & 19.61$\pm$0.22 & 26.18$\pm$0.36 & 31.12$\pm$0.18 & \textit{59.58$\pm$0.32} & \textbf{60.55$\pm$1.78}  &  \underline{59.67$\pm$0.11}      &  57.80$\pm$0.01 \\
                          & F1     & 41.28$\pm$2.43 & 60.96$\pm$0.22 & 64.25$\pm$0.22 & 63.21$\pm$0.12 & 43.53$\pm$0.42 & 57.14$\pm$0.17 & 61.82$\pm$0.13 & \textit{66.15$\pm$0.15} & \rm{66.16$\pm$0.64}  &  63.66$\pm$0.03      &  \textbf{66.54$\pm$0.01}\\ \hline
						  
\multirow{4}{*}{ACM}      & ACC    & 67.31$\pm$0.71 & 81.83$\pm$0.08 & 84.33$\pm$0.76 & 85.12$\pm$0.52 & 84.52$\pm$1.44 & 84.13$\pm$0.22 & 86.94$\pm$2.83 & 90.45$\pm$0.18          & \rm{90.59$\pm$0.15}  &  92.56$\pm$0.01      &  \textbf{92.61$\pm$0.03}\\
                          & NMI    & 32.44$\pm$0.46 & 49.30$\pm$0.16 & 54.54$\pm$1.51 & 56.61$\pm$1.16 & 55.38$\pm$1.92 & 53.20$\pm$0.52 & 56.18$\pm$4.15 & 68.31$\pm$0.25          & \rm{68.38$\pm$0.45}  &  73.27$\pm$0.03      &  \textbf{73.65$\pm$0.08}\\
                          & ARI    & 30.60$\pm$0.69 & 54.64$\pm$0.16 & 60.64$\pm$1.87 & 62.16$\pm$1.50 & 59.46$\pm$3.10 & 57.72$\pm$0.67 & 59.35$\pm$3.89 & 73.91$\pm$0.40          & \rm{74.20$\pm$0.38}  &  79.19$\pm$0.03      &  \textbf{79.36$\pm$0.07}\\
                          & F1     & 67.57$\pm$0.74 & 82.01$\pm$0.08 & 84.51$\pm$0.74 & 85.11$\pm$0.48 & 84.65$\pm$1.33 & 84.17$\pm$0.23 & 87.07$\pm$2.79 & 90.42$\pm$0.19          & \rm{90.58$\pm$0.17}  &  92.54$\pm$0.01      &  \textbf{92.59$\pm$0.02}\\ \hline
						  
\multirow{4}{*}{DBLP}     & ACC    & 38.65$\pm$0.65 & 51.43$\pm$0.35 & 58.16$\pm$0.56 & 60.31$\pm$0.62 & 61.21$\pm$1.22 & 58.59$\pm$0.06 & 62.05$\pm$0.48 & 68.05$\pm$1.81          & \rm{73.26$\pm$0.37}  &  77.67$\pm$0.14      &  \textbf{77.69$\pm$0.05}\\
                          & NMI    & 11.45$\pm$0.38 & 25.40$\pm$0.16 & 29.51$\pm$0.28 & 31.17$\pm$0.50 & 30.80$\pm$0.91 & 26.92$\pm$0.06 & 32.49$\pm$0.45 & 39.50$\pm$1.34          & \rm{39.68$\pm$0.42}  &  47.05$\pm$0.16      &  \textbf{47.12$\pm$0.06}\\
                          & ARI    & 6.97$\pm$0.39  & 12.21$\pm$0.43 & 23.92$\pm$0.39 & 25.37$\pm$0.60 & 22.02$\pm$1.40 & 17.92$\pm$0.07 & 21.03$\pm$0.52 & 39.15$\pm$2.01          & \rm{42.49$\pm$0.31}  &  \textbf{51.07$\pm$0.22}     &  50.22$\pm$0.07 \\
                          & F1     & 31.92$\pm$0.27 & 52.53$\pm$0.36 & 59.38$\pm$0.51 & 61.33$\pm$0.56 & 61.41$\pm$2.23 & 58.69$\pm$0.07 & 61.75$\pm$0.67 & 67.71$\pm$1.51          & \rm{72.80$\pm$0.56}  &  77.27$\pm$0.13      &  \textbf{77.49$\pm$0.05}\\ \hline
						  
\multirow{4}{*}{Citeseer} & ACC    & 39.32$\pm$3.17 & 57.08$\pm$0.13 & 55.89$\pm$0.20 & 60.49$\pm$1.42 & 61.35$\pm$0.80 & 60.97$\pm$0.36 & 64.54$\pm$1.39 & 65.96$\pm$0.31          & \rm{68.79$\pm$0.23}  &  73.19$\pm$0.06      &  \textbf{73.29$\pm$0.01}\\
                          & NMI    & 16.94$\pm$3.22 & 27.64$\pm$0.08 & 28.34$\pm$0.30 & 27.17$\pm$2.40 & 34.63$\pm$0.65 & 32.69$\pm$0.27 & 36.41$\pm$0.86 & 38.71$\pm$0.32          & \rm{41.54$\pm$0.30}  &  46.74$\pm$0.10      &  \textbf{46.92$\pm$0.02}\\
                          & ARI    & 13.43$\pm$3.02 & 29.31$\pm$0.14 & 28.12$\pm$0.36 & 25.70$\pm$2.65 & 33.55$\pm$1.18 & 33.13$\pm$0.53 & 37.78$\pm$1.24 & 40.17$\pm$0.43          & \rm{43.79$\pm$0.31}  &  50.01$\pm$0.12      &  \textbf{50.21$\pm$0.02}\\
                          & F1     & 36.08$\pm$3.53 & 53.80$\pm$0.11 & 52.62$\pm$0.17 & 61.62$\pm$1.39 & 57.36$\pm$0.82 & 57.70$\pm$0.49 & 62.20$\pm$1.32 & 63.62$\pm$0.24          & \rm{62.37$\pm$0.21}  &  63.34$\pm$0.04      &  \textbf{63.41$\pm$0.01} \\ \hline
\end{tabular}
}
\end{table*}

\subsection{Baseline Methods}
We compare with four types of methods, namely raw features \cite{hartigan1979algorithm}, deep clustering using features only \cite{hinton2006reducing,xie2016unsupervised,guo2017improved}, deep clustering using feature and attention learnt structure \cite{wang2019attributed, peng2021attention}, deep clustering using feature and GCN learnt structure \cite{kipf2016variational,bo2020structural}. Essentially our method, deep clustering using feature and structure learnt via loss, forms a distinct separate type. The list below summarizes the models:

\begin{itemize}
\item \textbf{K-means} \cite{hartigan1979algorithm} is a classical clustering method using raw data.
\item \textbf{AE} \cite{hinton2006reducing} applies K-means \cite{hartigan1979algorithm} to deep representations learned by an auto-encoder. 
\item \textbf{DEC} \cite{xie2016unsupervised}, clusters data in a jointly optimized feature space.
\item \textbf{IDEC} \cite{guo2017improved} enhances DEC by adding KL divergence based reconstruction loss
\item \textbf{GAE} \cite{kipf2016variational} combines convolution with  the AE
\item \textbf{DAEGC} \cite{wang2019attributed}, uses an attentional neighbour-wise strategy and clustering loss  
\item \textbf{SDCN} \cite{bo2020structural}, couples DEC and GCN via a fixed delivery operator and uses feature reconstruction
\item \textbf{AGCN} \cite{peng2021attention}, extends SDCN by adding an attention based delivery operator and uses multi scale information for cluster prediction. 
\end{itemize}

\textbf{Evaluation Metrics:}
Following \cite{bo2020structural, peng2021attention}, we use Accuracy (ACC), Normalized Mutual Information (NMI), Average Rand Index (ARI), and macro F1-score (F1) for evaluation. For each, larger values imply better clustering.

\subsection{Implementation} \label{Implementation}
\acrshort{MODEL} does not require an adjacency matrix for feed forward, and a once only computed reference is used during training. Note that this is not required for inference. An indirect benefit is that \acrshort{MODEL} can be trained in batches, facilitating scalability, without the need for full graph information as in most conventional GNNs \citep{kipf2016semi, bo2020structural, peng2021attention}. Batching can be implemented by randomly sampling $B$ nodes taking the corresponding adjacency information $\widehat{A} \in \mathbb{R} ^{B\times B}$, pre computed as in Equation~\ref{EQ_IC_DEF} for some $k$ hop depth of influence and the node features $\mathbf{X} \in R^{\mathbb{R} \times d}$. 

For fair comparison, we use the same $500-500-2000-10$ AE dimensions as in \cite{xie2016unsupervised,guo2017improved,bo2020structural, peng2021attention}. We use the same pre-training procedure as in \cite{bo2020structural, peng2021attention}, i.e. $30$ epochs; learning rate of $10^{-3}$ for USPS, HHAR, ACM, DBLP and $10^{-4}$ for REUT and CITE; batch size of $256$. We directly re-use the publicly available pre-trained AE from \cite{bo2020structural}.

For the training phase, for each data set, we first initialize the cluster centres using $K$-means. Unlike \cite{bo2020structural, peng2021attention}, where the best solution is taken from 20 initializations, we only do this once. We set $\beta=10$ for HHAR and 0.1 for others. Following all the compared methods, we repeat the \acrshort{MODEL} experiments 10 times with $200$ epochs and report the mean and standard deviation to prevent extreme cases. We directly cite the results from \cite{bo2020structural, peng2021attention} for other models.

For all experiments, we replicate the exact same training loops, including internal evaluation metric calls, when measuring performance for fair comparison. Our code will be made publicly available.

\begin{figure*}[h]
  \includegraphics[width=0.9\textwidth]{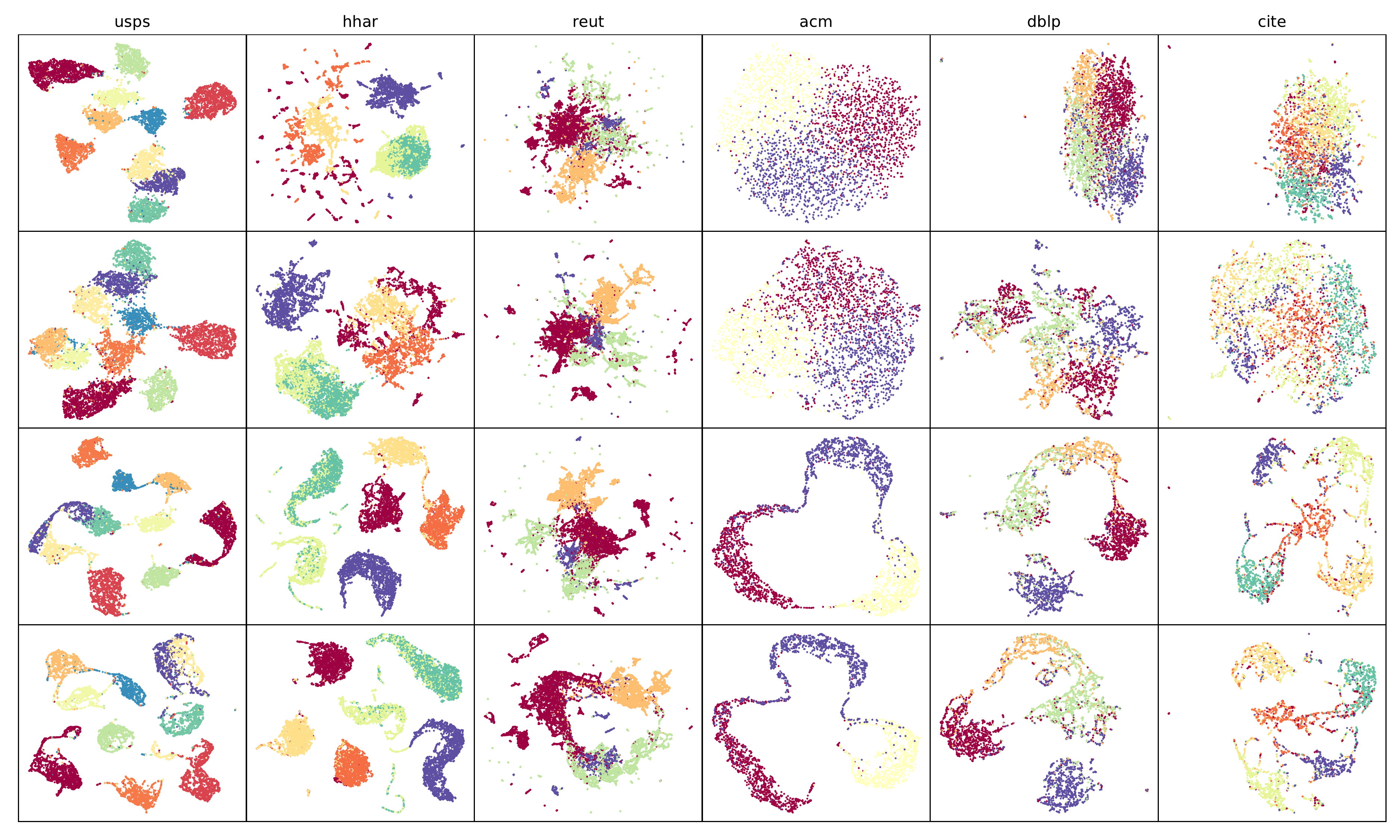}
  \caption{Visual comparison; top: embeddings from raw data, second row: embeddings from AE pre-training, third-row: embeddings from \acrshort{MODEL} and last-row: embeddings from \acrshort{MODEL}*. Colours represent ground truth groups.}
  \label{fig_Embeddings}
\end{figure*}

\subsection{Quantitative Results}
In Table~\ref{table_results}, we compare our results with state-of-the-art graph clustering methods \cite{bo2020structural, peng2021attention} following identical procedures. Our hyper parameters ($\alpha,K,\tau$) in dataset order are (1,4,0.5), (1,4,2.25), (3,3,1), (0.5,2,0.25), (0.5,1,0.25), (1,1,0.25) for \acrshort{MODEL} and (4,4,0.25), (1,3,2.25), (0.5,3,0.25), (1,1,0.25), (1,1,0.25), (1,1,0.25) for \acrshort{MODEL}*. Learning rate is $10^{-4}$ for CITE and $10^{-3}$ for others. We observe the following:
\begin{itemize}
\item For every metric, our methods \acrshort{MODEL} and \acrshort{MODEL}* achieves best ACC on all six data sets. Specifically, our approach achieves a significant improvement of 20\% on ARI and 18\% on NMI on DBLP. In CITE, we show improvements of 14\% and 12\% on ARI and NMI respectively. For both CITE and DBLP we improve ACC by 6.4\%  over prior state-of-the-art. Finally, for the USPS image dataset, our \acrshort{MODEL}* improves 5\% while \acrshort{MODEL} improves 2.4\% on ACC respectively and \acrshort{MODEL}* gains 8\% on ARI, 5.7\% on NMI and 5.1\% on F1.

\item These improvements arise from \acrshort{MODEL} being able to fuse multi-level node influences flexibly without needing to use rigid layers, and thus not requiring to make prior assumptions on the latent dependency structure of nodes. Further, all our model constraints are imposed softly; features are enforced with a soft reconstruction loss, structure is enforced with a soft contrastive loss and clustering is enforced with KL divergence for soft assignment. Thus, by seamlessly combining these soft constraints, we are able to handle diverse data modalities and characteristics effectively.

\item Generally we achieve better results on the natural graph datasets; ACM, DBLP and CITE. This can be expected as the constructed KNN based structures in the non-graph datasets may not correctly capture neighbour information. It also indicates the presence of more than simple similarity information in real-world graph structures, which can aid clustering and other downstream tasks. 

\item SDCN exceeds purely AE-based clustering methods (AE, DEC, IDEC) and purely GCN-based methods (GAE, VGAE, ARGA) by coupling AE and GCN models together, AGCN improves on SDCN by using attention based coupling, indicating the importance of a flexible or soft coupling between the feature and structure information. We completely eliminate any coupling, and replace it with a purely virtual guidance based supervisory linkage. Further, while AGCN uses multi-scale features, there is still the assumption of a latent structure that needs to be matched with convolutional layers. We solve the aforementioned drawbacks to achieve state-of-the-art on all six data sets.

\item We do not improve on NMI, ARI and F1 in REUT data set. This can be explained by the fact that its graph quality is poor \cite{peng2021attention}, resulting in poor clustering performance. Also, its class imbalance (4312:2403:2471:814) can result in high ACC if most points fall into the same large cluster. In contrast, for all natural real world graphs performance improvement of our method is significant.


\item Compared to \acrshort{MODEL}, \acrshort{MODEL}* achieves state-of-the-art by adopting influence based contrastive loss. However, as we show in Table~\ref{fig_performance}, \acrshort{MODEL}* also has better performance; overall 55\% reduction in training time and overall 81\% reduction on inference time over the next best model, AGCN.

\end{itemize}

\subsection{Qualitative Results}
In order to have a visual understanding of the embeddings, we visualize them via UAMP \citep{mcinnes2018umap} in Figure~\ref{fig_Embeddings}. Except for USPS, which is a distinct set of $0 \cdots 9$ handwritten digits, we see that all other data sets consists of quite indistinguishable clusters, as shown in the first row. In the second row, which uses the pre-trained AE based feature embeddings, the clusters are still not separable, especially for the real-world graph data sets. Incorporating structure via neighbour contrastivity and influence contrastivilty in row three (\acrshort{MODEL}) and four (\acrshort{MODEL}*) respectively, results in more cohesive and separable cluster formulation. 

\begin{figure*}
  \includegraphics[width=\textwidth]{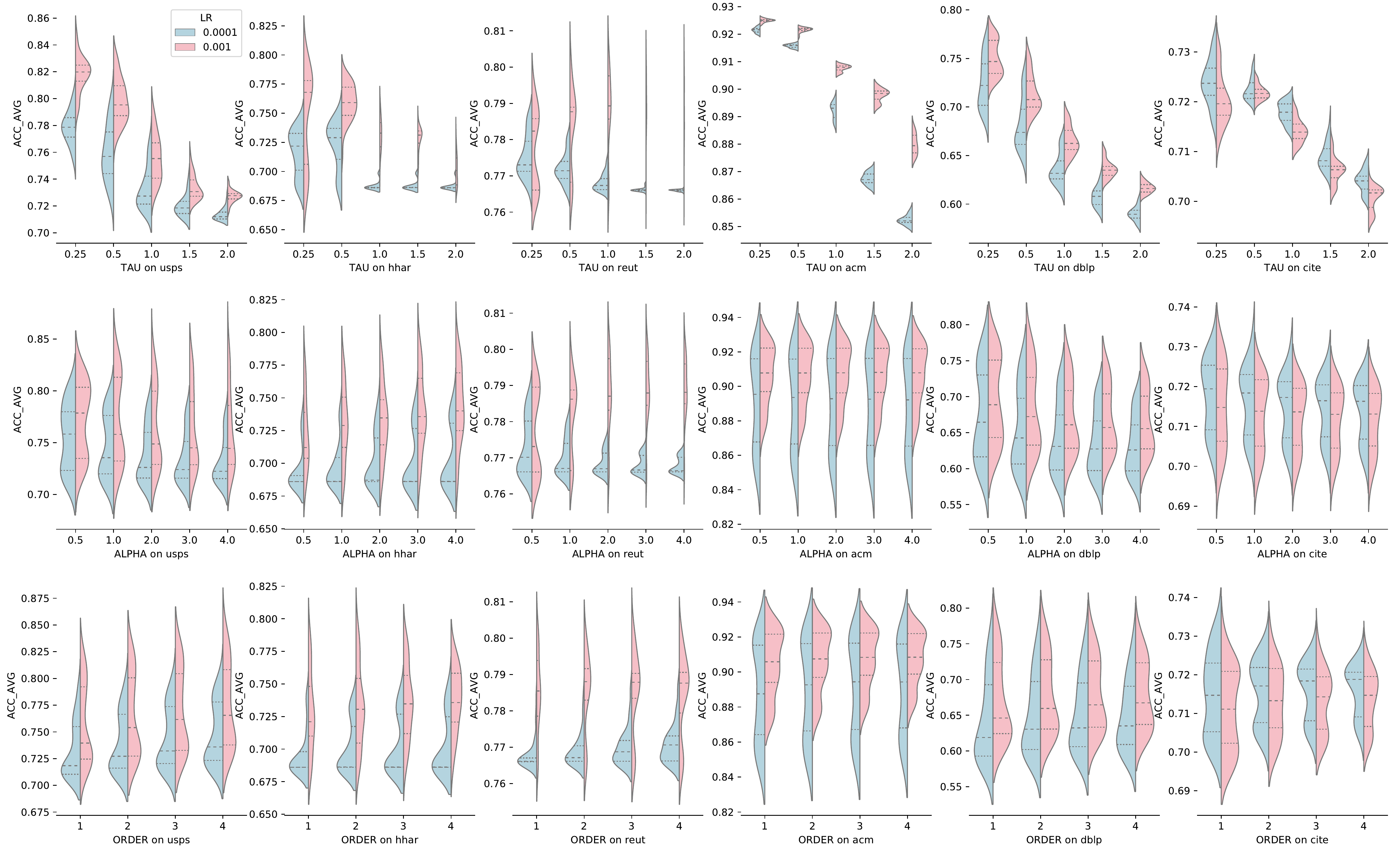}
  \caption{Ablation study on the hyper parameters. TAU=$\tau$, ALPHA=$\alpha$, ORDER=$R$ and LR denotes learning rate. A hyper parameter with higher and more condensed distribution represents its superiority over its counterpart. Best viewed in colour.}
  \label{fig_Hyperparam}
\end{figure*}

\subsection{Ablation study on Hyper-Parameters}
In order to have a deeper understanding of the effect of hyper parameters in \acrshort{MODEL}, we give an exhaustive analysis in Figure \ref{fig_Hyperparam}. Mainly we consider the hyper parameters of the contrastive loss from Equation~\ref{eq-LOSSES}: $\tau, \alpha, R$ and LR, the learning rate. We use other parameters similar to prior work \cite{bo2020structural, peng2021attention}. Overall, we observe that (1) $\alpha$ and $R$ are trivial hyper parameters as changes to these have relatively less effect on results. This shows \acrshort{MODEL} is robust to the mentioned hyper parameters. However, higher $R$ values are slightly better, which suggests that incorporating more neighbourhood influence is useful, which is the premise of \acrshort{MODEL}*. (2) As $\tau$ decreases accuracy is consistently improved on all data sets, particularly on DBLP and USPS. (3) A learning rate of 0.001 performs well relative to 0.0001, which is a recommended default setting.

\begin{figure}
    \centering \includegraphics[width=1.00\columnwidth]{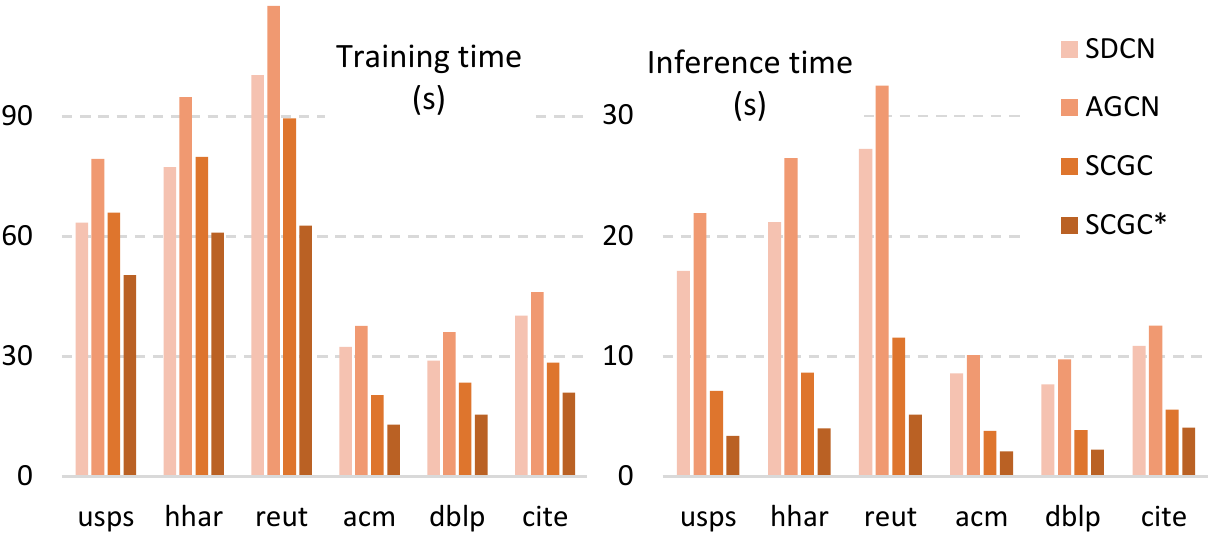}
    \caption{GPU time comparison against second best alternatives. Average GPU time (seconds) for 200 epochs on Google Colab, using the pytorch profiler.}
    \label{fig_performance}
\end{figure}

\subsection{Performance}
In Figure~\ref{fig_performance} we compare the GPU training and inference timings. Our model times also include the time taken for the cumulative influence computation. For all the data sets, \acrshort{MODEL} is superior without even the use of batching mentioned in Section~\ref{Implementation}. However, \acrshort{MODEL}* shows much more significant reductions in time, i.e., 27\% reduction in training time and 47\% reduction in inference time over \acrshort{MODEL}. Its MLP based design is lighter, the cluster centroids are better alternatives to the AE, and its IAC loss provides effective contrastive supervision. Additionally, in either model, inference does not need graph structure information, i.e., the adjacency matrix, which makes inference even faster. Also, for inference, time complexity is linearly related to the batch size. Further, inference can be trivially parallelized.
In comparison, \acrshort{MODEL}* averages 55\% reduction in training time and 81\% reduction in inference time over the second best model, AGCN. Thus, our models present strong efficiency advantages for resource constrained systems and edge-computing cases, particularly.

\subsection{Future work}
For comparison with prior work \cite{bo2020structural,peng2021attention}, we chose to use the same AE with $500-500-2000-10$ dimensions. However, further study is needed to determine an optimal AE to better handle the novel IAC loss, including diffident architectures (ex.: VAE \cite{kipf2016variational}, VQVAE), different layer choices (ex.: \cite{hu2021graph} uses Gelu activation, Layer normalization, and dropouts) and bottleneck sizes.
\acrshort{MODEL} has a time complexity of $\mathcal{O}(N^2)$, due to IAC loss, that can be reduced with batching or implementing a sparse version of the influence contrastive loss, which we leave for future work.

\section{Conclusion}\label{Conclusion}
This paper introduces \textbf{I}nfluence \textbf{A}ugmented \textbf{C}ontrastive (IAC), a novel influence-level contrastive loss for self-supervised learning of graph representations by adopting work from \cite{hu2021graph,zhong2021graph,you2020graph}. Its additive nature enables stronger inductive biases for generalization, without the necessity to approximate the latent dependency and relationship structure of complex graphs. Using IAC, our \acrshort{MODEL}* offers compelling advantages over traditional GNN based methods;   \acrshort{MODEL} readily accommodates local, long-term or any mixed combination of node dependencies naturally, supports batching, is efficient, has linear inference complexity and can be trivially parallelized. \acrshort{MODEL} achieves significant improvements over state-of-the-art (20\% on ARI 18\% on NMI for DBLP, 69\% reduction in training time for ACM, overall 55\% reduction in training time and overall, 81\% reduction on inference time).
By demonstrating novel possibilities between contrastive learning, clustering \cite{wu2020comprehensive,wang2019attributed,bo2020structural} and auto encoder models \cite{hinton2006reducing,xie2016unsupervised,guo2017improved} for effective graph clustering, we hope to provide inspiration and guidance for future model improvements in these fields and to address some of the limitations outlined in this paper.

\section{Acknowledgments}
The work was funded by the UQ RTP scholarship and supported by the Central Bank of Sri Lanka. Dedicated to Sugandi.

\bibliographystyle{ACM-Reference-Format}
\bibliography{main}
\end{document}